\documentclass[10pt,twocolumn,letterpaper]{article}

\usepackage{cvpr}              

\usepackage{graphicx}
\usepackage{amsmath}
\usepackage{amssymb}
\usepackage{booktabs}

\usepackage{epsfig}
\usepackage{graphicx}
\usepackage{amsmath}
\usepackage{amssymb}
\usepackage{multirow}
\usepackage{enumerate}
\usepackage{bm}
\usepackage{wrapfig}
\usepackage{epstopdf}
\usepackage{wrapfig}

\usepackage[pagebackref,breaklinks,colorlinks]{hyperref}

\usepackage[capitalize]{cleveref}
\crefname{section}{Sec.}{Secs.}
\Crefname{section}{Section}{Sections}
\Crefname{table}{eTable}{Tables}
\crefname{table}{Tab.}{Tabs.}

\def\cvprPaperID{536} 
\def\confName{CVPR}
\def\confYear{2022}

\begin{document}

\title{Surface Reconstruction from Point Clouds by Learning Predictive Context Priors}



\author{Baorui Ma$^{1}$, Yu-Shen Liu$^{1}$\thanks{The corresponding author is Yu-Shen Liu. This work was supported by National Key R$\&$D Program of China (2018YFB0505400, 2020YFF0304100), the National Natural Science Foundation of China (62072268), the National Natural Science Foundation (1813583) and in part by Tsinghua-Kuaishou Institute of Future Media Data.}, Matthias Zwicker$^2$, Zhizhong Han$^3$\\
$^1$School of Software, BNRist, Tsinghua University, Beijing, China\\
$^2$Department of Computer Science, University of Maryland, College Park, USA\\
$^3$Department of Computer Science, Wayne State University, Detroit, USA\\
{\tt\small mbr18@mails.tsinghua.edu.cn, liuyushen@tsinghua.edu.cn, zwicker@cs.umd.edu, h312h@wayne.edu}
}
\maketitle

\begin{abstract}
Surface reconstruction from point clouds is vital for 3D computer vision. State-of-the-art methods leverage large datasets to first learn local context priors that are represented as neural network-based signed distance functions (SDFs) with some parameters encoding the local contexts. To reconstruct a surface at a specific query location at inference time, these methods then match the local reconstruction target by searching for the best match in the local prior space (by optimizing the parameters encoding the local context) at the given query location.
However, this requires the local context prior to generalize to a wide variety of unseen target regions, which is hard to achieve.
To resolve this issue, we introduce \textit{Predictive Context Priors} by learning \textit{Predictive Queries} for each specific point cloud at inference time. 
Specifically, we first train a local context prior using a large point cloud dataset similar to previous techniques.
For surface reconstruction at inference time, however, we specialize the local context prior into our Predictive Context Prior by learning Predictive Queries, which predict adjusted spatial query locations as displacements of the original locations.
This leads to a global SDF that fits the specific point cloud the best.
Intuitively, the query prediction enables us to flexibly search the learned local context prior over the entire prior space, rather than being restricted to the fixed query locations, and this improves the generalizability.
Our method does not require ground truth signed distances, normals, or any additional procedure of signed distance fusion across overlapping regions. Our experimental results in surface reconstruction for single shapes or complex scenes show significant improvements over the state-of-the-art under widely used benchmarks. Code and data are available at 
\href{https://github.com/mabaorui/PredictableContextPrior}{https://github.com/mabaorui/PredictableContextPrior}.
\vspace{-0.3in}
\end{abstract}

\section{Introduction}
Surface reconstruction from 3D point clouds estimates continuous surfaces from 3D point clouds that can be captured by various 3D sensors. This is still a challenge even with the help of state-of-the-art deep learning models. A standard  strategy~\cite{gropp2020implicit,Atzmon_2020_CVPR,ErlerEtAl:Points2Surf:ECCV:2020} is to first learn a Signed Distance Function (SDF) from a point cloud~\cite{gropp2020implicit,Atzmon_2020_CVPR} or from ground truth signed distances~\cite{ErlerEtAl:Points2Surf:ECCV:2020} using a neural network, and then reconstruct a surface based on the learned SDF via marching cubes~\cite{Lorensen87marchingcubes}. If the SDF is trained to capture a global shape prior from a global 3D shape, however, it is hard to capture local geometry details.

As a remedy, state-of-the-art methods learn local SDFs from local regions~\cite{jiang2020lig,Tretschk2020PatchNets,DBLP:conf/eccv/ChabraLISSLN20}. The global shape is usually split into overlapping~\cite{jiang2020lig,Tretschk2020PatchNets} or non-overlapping~\cite{DBLP:conf/eccv/ChabraLISSLN20} parts, and the local region prior is learned as a local SDF that is represented by a neural network with some parameters encoding the geometry of local regions. The intuition behind this idea is that the local region prior will generalize to various unseen local reconstruction targets, and for surface reconstruction at inference time, its parameters can be optimized to match the reconstruction target at specific locations. However, the matching requires the learned local region prior to cover as many specific locations on target regions as possible, which dramatically limits the generalization ability of the learned local prior.


\begin{figure}[tb]
  \centering
   \includegraphics[width=\linewidth]{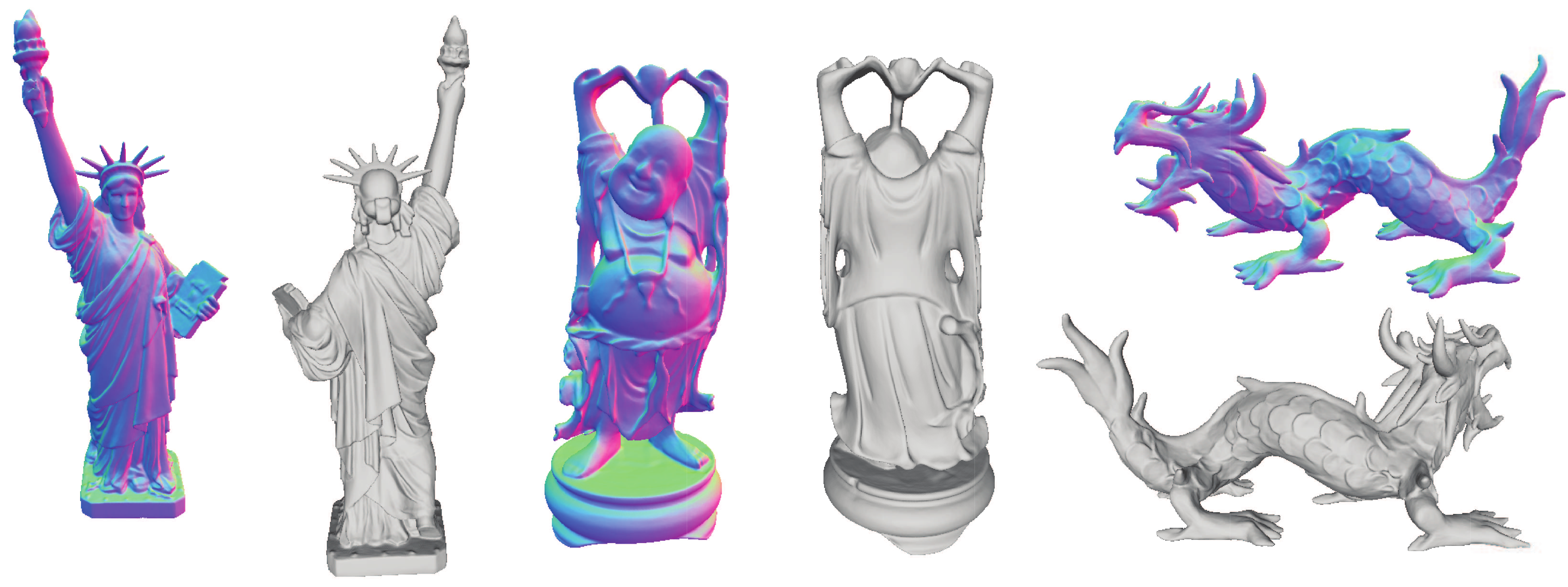}
  %
  %
  \vspace{-0.3in}
\caption{\label{fig:Teaser}We reconstruct highly accurate surfaces from 300K points. Front views of each shape are shown with normal maps. The comparison with Poisson is in supplemental materials.}
 \vspace{-0.28in}
\end{figure}

To resolve this issue, we propose to learn SDFs as a Predictive Context Prior for highly accurate surface reconstruction from point clouds, as shown in Fig.~\ref{fig:Teaser}. Specifically, we first train a neural network to represent local SDFs of local regions across a large dataset of point clouds. This aims to capture a local context prior in a local coordinate system, similar as in previous work. Our main contribution is that during surface reconstruction at inference time, we specialize the pre-trained local context prior into a \textit{Predictive Context Prior} for a specific point cloud by learning \textit{Predictive Queries}. More concretely, Predictive Queries learn to predict query locations for the pre-trained local SDF from queries given in the global coordinate system of the specific point cloud. This is achieved by sampling a set of queries in the global coordinate system, and learning to predict queries for the local SDF to minimize surface reconstruction error.

Intuitively, learning to predict query locations for the local SDF allows us to more flexibly match the learned local context prior to best fit the given point cloud during surface reconstruction. This further leads to a global SDF for the specific point cloud.
Our method does not require ground truth signed distances, normals in the training, or any additional post processing. Our contributions are as follows:


\begin{enumerate}[i)]
\item We introduce a neural network architecture using Predictive Context Priors to learn SDFs for surface reconstruction from point clouds. Predictive Context Priors are implemented by learning Predictive Queries.
\item We demonstrate that Predictive Queries allow us to improve generalizability of a pre-trained local context prior, which improves reconstruction accuracy.
\item We report state-of-the-art results in surface reconstruction for single shapes or complex scenes under widely used benchmarks.
\end{enumerate}

\section{Related Work}
Surface reconstruction has been studied for decades. Classic methods~\cite{journals/tog/KazhdanH13,Lorensen87marchingcubes,817351TVCG} do not leverage any prior learned from large scale datasets. With the development of deep learning, data driven strategies~\cite{Zhu2021NICESLAM,ruckert2021adop,Park_2019_CVPR,MeschederNetworks,mildenhall2020nerf,Zhizhong2018seq,Zhizhong2019seq,3D2SeqViews19,wenxin_2020_CVPR,seqxy2seqzeccv2020paper,Zhizhong2018VIP,Zhizhong2020icml,Han2019ShapeCaptionerGCacmmm,zhizhongiccv2021finepoints,MAPVAE19,p2seq18,hutaoaaai2020,wenxin_2021a_CVPR,wenxin_2021b_CVPR,Jiang2019SDFDiffDRcvpr,zhizhongiccv2021completing,tianyangcvpr2022,wenxincvpr2022,jain2021dreamfields,text2mesh,yu_and_fridovichkeil2021plenoxels,mueller2022instant} can learn effective priors from datasets to improve the surface reconstruction accuracy. We will focus on reviewing the studies of deep learning based methods.

\noindent\textbf{Deep Learning based Surface Reconstruction. }The state-of-the-art methods mainly represent the reconstruction target as an implicit function~\cite{MeschederNetworks,Park_2019_CVPR,Genova_2020_CVPR,Tretschk2020PatchNets,chen2018implicit_decoder,seqxy2seqzeccv2020paper,Jiang2019SDFDiffDRcvpr,ICML21Effectiveness,sitzmann2019siren,rematasICML21,tancik2020fourfeat,Oechsle2021ICCV}, due to advantages of SDFs or occupancy fields over other representations in representing high resolution shapes with arbitrary topology. To reveal more detailed geometry, one strategy is to leverage more latent codes~\cite{DBLP:journals/corr/abs-2105-02788,takikawa2021nglod} to capture local shape priors as SDFs~\cite{jiang2020lig,DBLP:conf/eccv/ChabraLISSLN20,9320319Lombardi}. This requires to split the point cloud into different voxels, and then represent the points in each voxel as a latent code that is either extracted by a neural network~\cite{jiang2020lig,9320319Lombardi} or learned in an auto-decoding manner~\cite{jiang2020lig,DBLP:conf/eccv/ChabraLISSLN20}. These methods need normals for each point to produce signed distances as supervision in the optimization. Given the ground truth signed distances, Points2Surf~\cite{ErlerEtAl:Points2Surf:ECCV:2020} encodes points sampled in a local patch and on the whole point cloud as a shape prior, while DeepMLS~\cite{Liu2021MLS} learns to produce oriented points to approximate SDFs. Similarly, PatchNet~\cite{Tretschk2020PatchNets} learns local SDFs to represent patches with explicit control over positions, orientations, and scales. Neural-pull~\cite{Zhizhong2021icml} introduced a new way of learning SDFs by pulling nearby space onto the surface, which is achieved by predicting the SDFs and its gradient using the network. This removes the requirement of ground truth normals or signed distances. A similar idea is introduced to learn unsigned distances~\cite{chibane2020neural}, but requires to move dense sampling with additional directions to form the surface. Moreover, other novel ways for surface reconstruction have been proposed, such as a differentiable formulation of Poisson solver~\cite{nipspoisson21}, point convolution~\cite{pococvpr2022} and part retrieval~\cite{siddiqui2021retrievalfuse}.

Other information is also leveraged to learn implicit functions~\cite{Mi_2020_CVPR,Peng2020ECCV,yifan2020isopoints}. Occupancy is used to capture a prior at a global level~\cite{Peng2020ECCV,jia2020learning} or a local level~\cite{Mi_2020_CVPR}. Iso-points~\cite{yifan2020isopoints} tried to impose
geometry-aware sampling and regularization in the learning. Moreover, implicit functions can also be learned from point clouds with additional constraints, such as geometric regularization~\cite{gropp2020implicit}, sign agnostic learning with a specially designed loss function~\cite{Atzmon_2020_CVPR}, sign agnostic learning with local surface self-similarities and post sign processing~\cite{zhao2020signagnostic,tang2021sign}, constraints on gradients~\cite{atzmon2020sald} or a divergence penalty~\cite{DBLP:journals/corr/abs-2106-10811}.

From a meshing perspective, surfaces can also be reconstructed by generating local connectivity with intrinsic-extrinsic metrics~\cite{liu2020meshing}, Delaunay triangulation of point clouds~\cite{luo2021deepdt} or inheriting connectivity from an initial mesh~\cite{Hanocka2020p2m}. With local chart parameterizations in neural networks, a local point cloud is reconstructed via fitting using the Wasserstein distance as a measure of approximation~\cite{Williams_2019_CVPR}.

\noindent\textbf{Deep Shape Prior. }Beside the priors reviewed above, shape priors can also be captured by parameters in neural networks in shape reconstruction~\cite{yang2020deep,DBLP:conf/iccv/WallaceH19,NIPS2019_SELFREC,InsafutdinovD18,bednarik2020,Groueix_2018_CVPR,Zhizhong2020icml,Gadelha2019,MAPVAE19,Zhizhong2018VIP}, segmentation~\cite{nipspoint17,cvprpoint2017,p2seq18}, and completion~\cite{WuZZZFT18,Yuan-2018-110219,wenxin_2020_CVPR,hutaoaaai2020,Hu2019Render4CompletionSM}. Deep manifold prior~\cite{gadelha2020deep} was introduced to reconstruct 3D shapes starting from random initializations.

\section{Method}
\begin{figure}[tb]
  \centering
   \includegraphics[width=\linewidth]{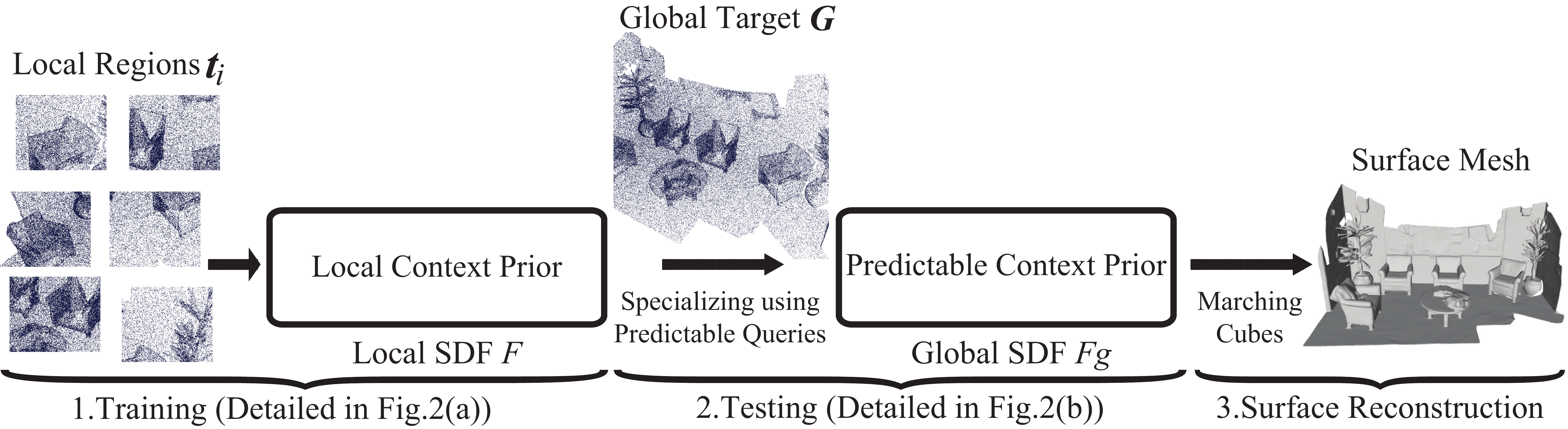}
  %
  %
   \vspace{-0.3in}
\caption{\label{fig:OverviewGeneral}Overview of our method.}
 \vspace{-0.28in}
\end{figure}

\noindent\textbf{Overview. }We provide an overview of our method in Fig.~\ref{fig:OverviewGeneral}. We aim to reconstruct a surface mesh for a 3D point cloud $\bm{G}$. Our method consists of the following three stages.

\begin{figure}[tb]
  \centering
   \includegraphics[width=\linewidth]{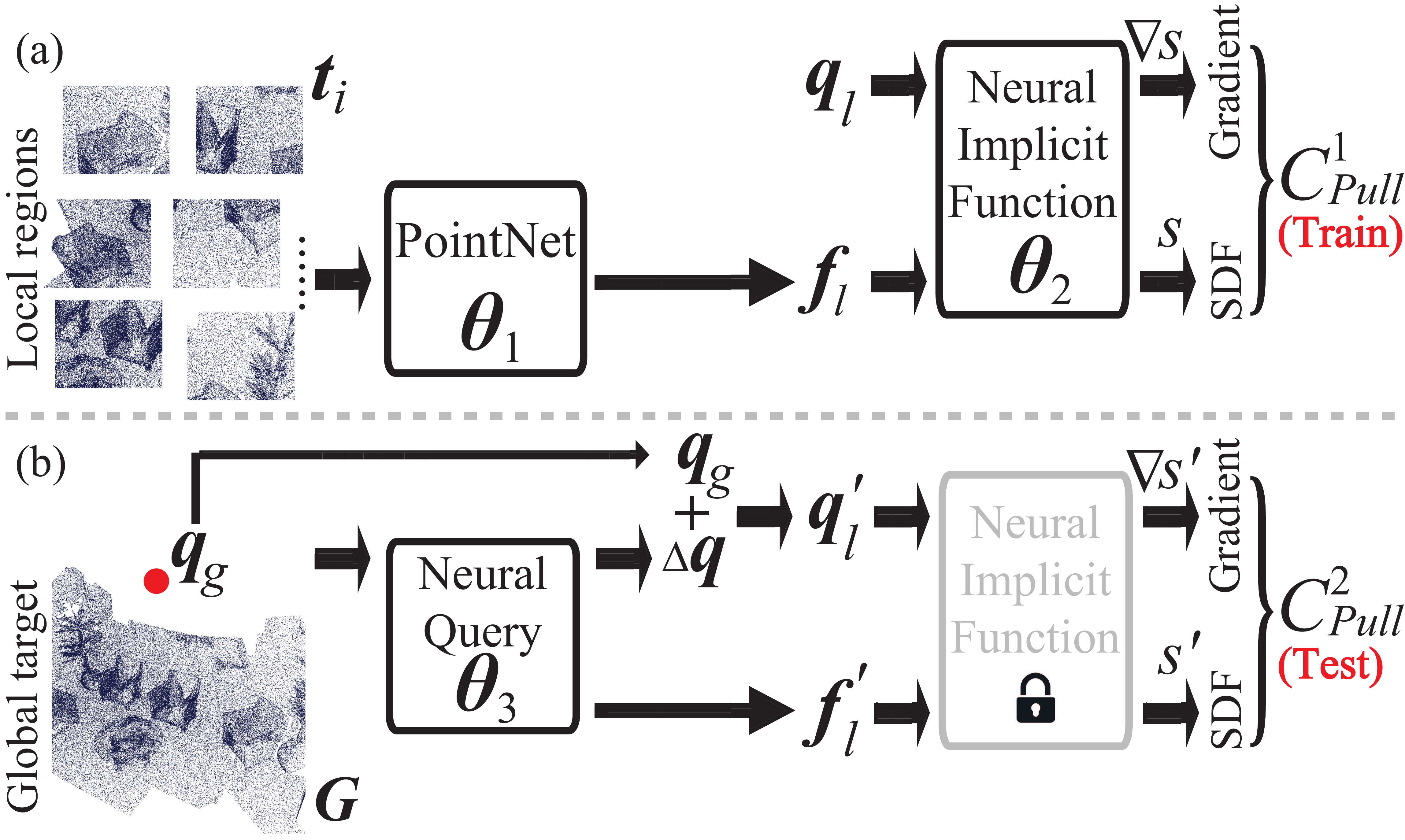}
  %
  %
   \vspace{-0.3in}
\caption{\label{fig:Overview}We learn local context prior from each local region $\bm{t}_i$ as a local SDF $F$ in (a). We learn a predictive context prior to reconstruct $\bm{G}$ by predicting queries $\bm{q}_l'$ associated with conditions $\bm{f}_l'$ for $F$ in (b).}
 \vspace{-0.28in}
\end{figure}

\noindent 1. During training, we start by learning a local context prior as a local SDF $F$ by training a neural implicit network under a local region set $\mathbf{T}=\{\bm{t}_i,i\in[1,I]\}$. As shown in Fig.~\ref{fig:Overview} (a), the neural implicit network learns $F$ as a mapping from a query point $\bm{q}_l$ with its corresponding condition $\bm{f}_l$ to a signed distance $s$ in a local coordinate system.

\noindent 2. For surface reconstruction at test time, we specialize the local context prior into a predictive context prior for a specific point cloud $\bm{G}$ by learning predictive queries. The predictive context prior leads to a global SDF $F_g$ to fit the point cloud $\bm{G}$. Specifically, the learned local context prior is represented by the fixed parameters in the neural implicit network, as shown in Fig.~\ref{fig:Overview} (b).
We train an additional neural query network to predict queries $\bm{q}_l'$ and their conditions $\bm{f}_l'$ to explore the learned local context prior, which significantly improves the generalization ability of the learned local prior by enabling us to flexibly search the whole learned prior space.

\noindent 3. Finally, we leverage the global SDF $F_g$ to reconstruct the surface of $\bm{G}$ using the marching cubes algorithm~\cite{Lorensen87marchingcubes}.


\noindent\textbf{Local Context Prior. }During training shown in Fig.~\ref{fig:Overview} (a), we learn a local context prior from local point clouds $\bm{t}_i\in\mathbf{T}$ as a local SDF $F$ in a local coordinate system. We learn the context around each point of $\bm{t}_i$ using a neural implicit network. For a local point cloud $\bm{t}_i$, we normalize all points of $\bm{t}_i$ by centering them to the origin, and then linearly scaling them to fit the longest edge of the bounding box of $\bm{t}_i$ into a range of $[-0.5,0.5]$. This normalization makes the local context on different $\bm{t}_i$ comparable to each other.

We leverage PointNet~\cite{cvprpoint2017} to extract the feature $\bm{f}_l$ of each local region $\bm{t}_i$. $\bm{f}_l$ is regarded as a condition of any query point $\bm{q}_l$ sampled near $\bm{t}_i$ when training the neural implicit network as a local SDF $F$, such that the signed distance $s$ at the location of $\bm{q}_l$ is,

\begin{equation}
\label{eq:s}
\begin{aligned}
s=F(\bm{q}_l,\bm{f}_l).
\end{aligned}
\end{equation}

To remove the requirement of ground truth signed distance values or normals in training, we minimize a pulling cost introduced in Neural-pull~\cite{Zhizhong2021icml} to train the local SDF $F$. We simultaneously optimize the parameters of $\bm{\theta}_1$ in PointNet and $\bm{\theta}_2$ in the neural implicit network. The intuition of the pulling cost is to pull a query $\bm{q}_l$ using the predicted signed distance $s$ to its nearest neighbor $nn(\bm{q}_l)$ on region $\bm{t}_i$ along the direction of the gradient $\nabla s=\partial F/\partial \bm{q}_l$ at $\bm{q}_l$. Our objective function during training is to minimize the pulling cost $C_{Pull}^1$, where $nn(\bm{q}_l)\in\bm{t}_i$,

\begin{equation}
\label{eq:o}
\begin{aligned}
\min_{\bm{\theta}_1,\bm{\theta}_2}\|nn(\bm{q}_l)-(\bm{q}_l-s\times \nabla s/\|\nabla s\|_2)\|_2.
\end{aligned}
\end{equation}


\noindent\textbf{Predictive Context Prior. }For surface reconstruction at test time, we first specialize the learned local context prior into a predictive context prior for a specific point cloud $\bm{G}$. Point cloud $\bm{G}$ is located in a global coordinate system without normalization. We train an additional neural query network with parameters of $\bm{\theta}_3$ specially for $\bm{G}$, where we keep the neural network parameters $\bm{\theta}_2$ representing the learned local context prior fixed. This leads to a global SDF $F_g$ that captures the predictive context prior which we use to reconstruct the surface of $\bm{G}$.

The neural query network learns to generate predictive queries, that is, to transform a query point $\bm{q}_g$ around $\bm{G}$ in a global coordinate system into a point $\bm{q}_l'$ in the local coordinate system that the learned local context prior covers. In addition, the neural query network also predicts the condition $\bm{f}_l'$ of predictive query $\bm{q}_l'$. Here, we are inspired by the idea of ResNet~\cite{DBLPHeZRS16}, and predict the shift $\Delta \bm{q}$ from $\bm{q}_g$ to $\bm{q}_l'$,

\begin{equation}
\label{eq:s1}
\begin{aligned}
\bm{q}_l'=\bm{q}_g+\Delta \bm{q}.
\end{aligned}
\end{equation}

The intuition behind the neural query network is to train a network specific to point cloud $\bm{G}$ that is able to manipulate the queries for the learned local context prior. This prediction is equivalent to flexibly searching for correct information from the learned local SDF $F$, and then combining them together to fit the point cloud $\bm{G}$. This leads to a global SDF $F_g$ that predicts the signed distance $s'$ at a query location of $\bm{q}_g$ with a condition of $\bm{G}$,

\begin{equation}
\label{eq:s}
\begin{aligned}
s'=F_g(\bm{q}_g,\bm{G})=F(\bm{q}_l',\bm{f}_l').
\end{aligned}
\end{equation}

Similar to Eq.~(\ref{eq:o}) in training, we further optimize the parameters $\bm{\theta}_3$ of the neural query network to pull the query $\bm{q}_g$ in the global coordinate system to its nearest neighbor $nn(\bm{q}_g)$ on point cloud $\bm{G}$. We leverage the learned local SDF to produce the gradient, $\nabla s'=\partial F/\partial \bm{q}_l'$. So, our objective function during testing is to minimize a pulling cost $C_{Pull}^2$ below, where $nn(\bm{q}_g)\in\bm{G}$,

\begin{equation}
\label{eq:o1}
\begin{aligned}
\min_{\bm{\theta}_3}\|nn(\bm{q}_g)-(\bm{q}_g-s'\times \nabla s'/\|\nabla s'\|_2)\|_2.
\end{aligned}
\end{equation}

\noindent\textbf{Reconstruction. }After we learn parameters $\bm{\theta}_3$ in the neural query network using Eq.~(\ref{eq:o1}), we keep $\bm{\theta}_3$ and the parameters $\bm{\theta}_2$ in the neural implicit network fixed to produce the global SDF $F_g$ for point cloud $\bm{G}$, which is then used to reconstruct the surface using marching cubes~\cite{Lorensen87marchingcubes}.

\noindent\textbf{Optimization. }We conduct the optimization of Eq.~(\ref{eq:o})(training) and Eq.~(\ref{eq:o1})(inference) using a similar procedure. For each point $\bm{p}$ on $\bm{t}_i$ or $\bm{G}$, we randomly sample 40 queries $\bm{q}_l$ around $\bm{p}$. Due to the difference numbers of points on each local region $\bm{t}_i\in\mathbf{T}$, we randomly select 2000 $\bm{q}_l$ around $\bm{t}_i$, and regard their nearest neighbors $\{nn(\bm{q}_l)\}$ on $\bm{t}_i$ as the input to PointNet in each training epoch, where the randomness makes the local context prior more robust to noise. We perform this optimization in an overfitting manner, either on each single $\bm{G}$ or multiple point clouds with one-hot vectors as the condition of each $\bm{q}_g$.



\begin{figure}[tb]
  \centering
   \includegraphics[width=\linewidth]{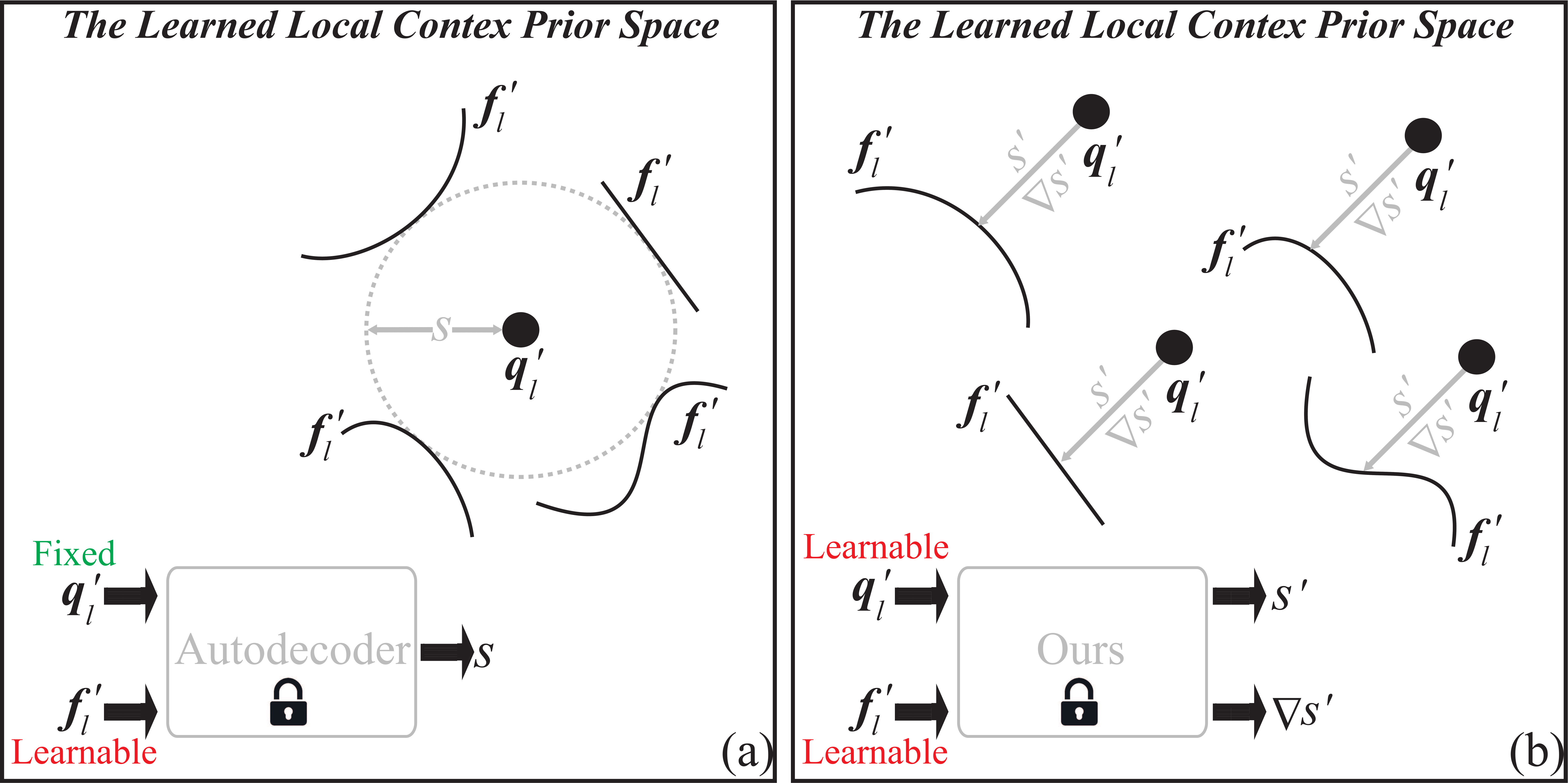}
  %
  %
   \vspace{-0.3in}
\caption{\label{fig:Intuition}Current methods can merely search similar regions at a specific location $\bm{q}_l'$ in the learned prior space by optimizing condition $\bm{f}_l'$. While we can search over the whole prior space by simultaneously optimizing $\bm{q}_l'$ and $\bm{f}_l'$.}
 \vspace{-0.28in}
\end{figure}

\noindent\textbf{Intuition and Advantages. }The intuition of our predictive context prior is to leverage information at different locations queried from the learned prior in local SDF $F$ to form a global SDF $F_g$, which significantly improves the generalization ability of the learned local context prior.

A widely used strategy to explore the local context prior is to use an autodecoder~\cite{Park_2019_CVPR,jiang2020lig,Zhizhong2018VIP}, as shown in Fig.~\ref{fig:Intuition} (a). The autoencoder aims to optimize a learnable condition $\bm{f}_l'$ for a fixed location $\bm{q}_l'$ to minimize the signed distance error compared to the ground truth $s$. This optimization is equivalent to finding a local region whose feature is $\bm{f}_l'$ with a nearest distance to the location $\bm{q}_l'$ as $s$ in the space covered by the learned local context prior. Since the specific location $\bm{q}_l'$ is fixed, the performance is only guaranteed when a qualified local region represented by $\bm{f}_l'$ has been seen during the learning of the local context prior, which is hard to generalize for various unseen regions during test.

Differently, without requiring the ground truth signed distances, our method aims to find a similar way of pulling a location $\bm{q}_l'$ to a local region represented by $\bm{f}_l'$, where the pulling is implemented by a signed distance prediction $s'$ and its gradient $\nabla s'$. As shown in Fig.~\ref{fig:Intuition} (b), rather than searching (optimizing $\bm{f}_l'$) at a fixed location $\bm{q}_l'$ like an autodecoder, our neural query network can adjust query locations, which makes it possible to search a similar pulling way across the whole space covered by the learned local context prior.

Obviously, our advantage is the ability of transforming the searching at a specific location into anywhere across the learned context prior. This advantage not only significantly improves the generalization ability of the learned prior, but also dramatically reduces the requirement of the local regions used to learn the local context prior, since it is easy to observe various ways of pulling points to the surface around arbitrary local regions during training.

\begin{figure}[tb]
  \centering
   \includegraphics[width=\linewidth]{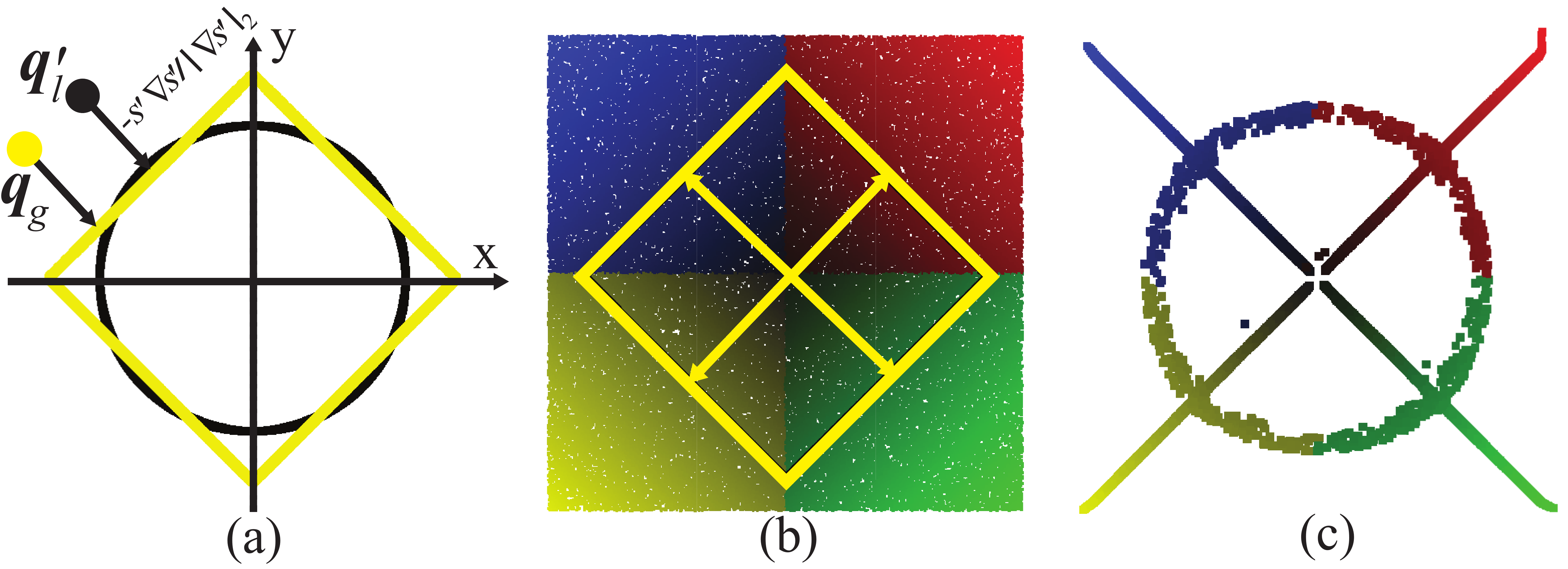}
  %
  %
   \vspace{-0.3in}
\caption{\label{fig:Intuition1}Demonstration of predicted queries $\bm{q}_l'$ (c) from queries $\bm{q}_g$ (b). Optimization is visualized in our video supplementary.}
 \vspace{-0.28in}
\end{figure}

We further demonstrate our advantages in a 2D case. We aim to estimate an SDF $F_g$ of a yellow square in Fig.~\ref{fig:Intuition1} (a) from an SDF $F$ learned from a black circle in Fig.~\ref{fig:Intuition1} (a) during training. During testing, we sample query points $\bm{q}_g$ in the space occupied by the yellow square, as shown by the dense points in color in Fig.~\ref{fig:Intuition1} (b). After the optimization using Eq.~(\ref{eq:o1}), we visualize the predicted queries $\bm{q}_l'$ obtained from Eq.~(\ref{eq:s1}) in Fig.~\ref{fig:Intuition1} (c), where the color of each $\bm{q}_l'$ is the same as the corresponding $\bm{q}_g$. The color correspondence demonstrates that the $\bm{q}_l'$ appearing on the radius with an angle of $45^{\circ}$ to the x-axis can provide the expected $s'$ and $\nabla s'$ to get the corresponding $\bm{q}_g$ pulled on the square, i.e., the appropriate signed distances $s'$ and gradients $\nabla s'$ with $45^{\circ}$ to the x-axis (shown by black arrows in Fig.~\ref{fig:Intuition1} (a) and yellow arrows in Fig.~\ref{fig:Intuition1} (b)), since $\bm{q}_l'$ gets pulled to the circle using the same $s'$ and $\nabla s'$, while the $\bm{q}_l'$ appearing on the circle correspond to $\bm{q}_g$ on the square, both of which have 0 signed distances. This example shows that our method can flexibly search over the whole learned local context prior and easily find the correct prior information. This significantly increases our generalization ability.

\section{Experiments, Analysis, and Discussion}

\subsection{Setup}
\noindent\textbf{Implementation details. }To predict a signed distance value $s$, we use an OccNet~\cite{MeschederNetworks} without activation functions in the last layer. The condition $\bm{f}_l'$ or $\bm{f}_l$ is 512 dimensional. The neural query network is a feed-forward network with 8 layers, where each one of the first 7 layers has 512 nodes with ReLU activation functions while the last layer has 515 nodes with linear activation functions to predict the 512 dimensional condition $\bm{f}_l'$ and 3 dimensional query $\bm{q}_l'$.

We separate each 3D shape or scene in the training set under each benchmark into a $6^3$ grid according to its bounding box, where the points located in each grid form a local region $\bm{t}_i$ in $\mathbf{T}$. In addition, we use the same method as Neural-pull~\cite{Zhizhong2021icml} to sample 40 queries $\bm{q}_l$ or $\bm{q}_g$ around each point $\bm{p}$ on $\bm{t}_i$ or $\bm{G}$, respectively, where a Gaussian function $\mathcal{N}(\bm{p},\sigma^2)$ is used to calculate the sampling probability, and $\sigma^2$ is the $50$-th nearest neighbor of $\bm{p}$ on $\bm{t}_i$ or $\bm{G}$.

\noindent\textbf{Dataset. }In surface reconstruction for 3D shapes, we evaluate our method under three datasets including ABC~\cite{Koch_2019_CVPR}, FAMOUS~\cite{ErlerEtAl:Points2Surf:ECCV:2020}, and a subset of ShapeNet~\cite{shapenet2015}. In surface reconstruction for scenes, we report our results under two datasets including 3D Scene~\cite{DBLP:journals/tog/ZhouK13} and SceneNet~\cite{7780811Handa}. Under ShapeNet and ABC, we leverage marching cubes on a $128^3$ grid to reconstruct meshes, while using a $512^3$ grid under FAMOUS, 3D Scene and SceneNet.

\noindent\textbf{Metrics. }
Under the ABC and FAMOUS datasets, we randomly sample $1\times10^4$ points on the reconstructed mesh to compare with the input point clouds using L2-CD which keeps the same as the setting in Points2Surf~\cite{ErlerEtAl:Points2Surf:ECCV:2020}. We also follow MeshingPoint~\cite{liu2020meshing} to report our results under ShapeNet~\cite{shapenet2015} in terms of L1-CD, Normal Consistency (NC)~\cite{MeschederNetworks}, and F-score~\cite{Tatarchenko_2019_CVPR} to evaluate the reconstruction performance, where we evaluate the distance between the $1\times10^5$ points sampled on the reconstructed shape and the $1\times10^5$ ground truth points released by OccNet~\cite{MeschederNetworks}. The threshold $\mu$ in F-score calculation is 0.002 which is the same as MeshingPoint~\cite{liu2020meshing} and Neural-pull~\cite{Zhizhong2021icml}.

Under the 3D Scene~\cite{DBLP:journals/tog/ZhouK13} dataset, we follow DeepLS~\cite{DBLP:conf/eccv/ChabraLISSLN20} to report the error between reconstructed meshes and the ground truth mesh. The error with a unit of mm is the average distance from each reconstructed vertex to its nearest triangle on the ground truth mesh. We also produce L1-CD, L2-CD, and normal consistency to compare with others.

Under the SceneNet~\cite{7780811Handa} dataset, we follow LIG~\cite{jiang2020lig} to report L1-CD, Normal Consistency (NC)~\cite{MeschederNetworks}, and F-score~\cite{Tatarchenko_2019_CVPR} under different sampling densities on the reconstructed meshes, such as 20, 100, 500
and 1000 points per $m^2$, where the threshold $\mu_s$ in F-score calculation is 0.025 m which is the same as LIG~\cite{jiang2020lig}.

\subsection{Surface Reconstruction for Single Shapes}
\noindent\textbf{Evaluation under ShapeNet. }We first report our numerical comparison under the ShapeNet subset by comparing with the non data-driven and the latest data-driven methods in terms of L2-CD in Tab.~\ref{table:t10}, normal consistency in Tab.~\ref{table:t11}, and F-score with a threshold of $\mu$ in Tab.~\ref{table:st12}, and $2\mu$ in Tab.~\ref{table:t13}. The compared methods include Poisson Surface Reconstruction (PSR)~\cite{journals/tog/KazhdanH13}, Ball-Pivoting algorithm (BPA)~\cite{817351TVCG}, AtlasNet (ATLAS)~\cite{Groueix_2018_CVPR}, Deep Geometric Prior (DGP)~\cite{Williams_2019_CVPR}, Deep Marching Cube (DMC)~\cite{Liao2018CVPR}, DeepSDF (DSDF)~\cite{Park_2019_CVPR}, MeshP~\cite{liu2020meshing}, Neural Unsigned Distance (NUD)~\cite{chibane2020neural}, SALD~\cite{atzmon2020sald}, Local Implicit Grid (LIG)~\cite{jiang2020lig}, IMNET~\cite{chen2018implicit_decoder}, and Neural-Pull(NP)~\cite{Zhizhong2021icml}.

The reconstruction accuracy in Tab.~\ref{table:t10} demonstrates that our method reveals the most accurate surface from point clouds even under some challenging classes, such as Lamp, Chair, and Table. Although we achieve comparable normal consistency to MeshP in Tab.~\ref{table:t11}, we do not require dense and clean point clouds as MeshP. In addition, our method outperforms all implicit function based methods including DSDF~\cite{Park_2019_CVPR}, NUD~\cite{chibane2020neural}, SALD~\cite{atzmon2020sald}, LIG~\cite{jiang2020lig}, IMNET~\cite{chen2018implicit_decoder}, NP~\cite{Zhizhong2021icml} in Tab.~\ref{table:t10} and ~\ref{table:t11}, which justifies our ability of leveraging prior information in a more effective way.

\begin{table}[tb]
\centering
\resizebox{\linewidth}{!}{
    \begin{tabular}{c|c|c|c|c|c|c|c|c|c|c|c|c}  
     \hline
        Class& PSR& DMC & BPA & ATLAS &DMC&DSDF& DGP &MeshP&NUD&SALD&NP&Ours \\  
     \hline
        Display& 0.273& 0.269 & 0.093 & 1.094 &0.662&0.317& 0.293 & 0.069 & 0.077 &-& 0.039&\textbf{0.0087}\\
        Lamp &0.227&0.244&0.060&1.988&3.377&0.955&0.167&0.053&0.075&0.071&0.080&\textbf{0.0380}\\
        Airplane&0.217&0.171&0.059&1.011&2.205&1.043&0.200&0.049&0.076&0.054&0.008&\textbf{0.0065}\\
        Cabinet&0.363&0.373&0.292&1.661&0.766&0.921&0.237&0.112&0.041&-&0.026&\textbf{0.0153}\\
        Vessel&0.254&0.228&0.078&0.997&2.487&1.254&0.199&0.061&0.079&-&0.022&\textbf{0.0079}\\
        Table&0.383&0.375&0.120&1.311&1.128&0.660&0.333&0.076&0.067&0.066&0.060&\textbf{0.0131}\\
        Chair&0.293&0.283&0.099&1.575&1.047&0.483&0.219&0.071&0.063&0.061&0.054&\textbf{0.0110}\\
        Sofa&0.276&0.266&0.124&1.307&0.763&0.496&0.174&0.080&0.071&0.058&0.012&\textbf{0.0086}\\
     \hline
     Mean&0.286&0.276&0.116&1.368&1.554&0.766&0.228&0.071&0.069&0.062&0.038&\textbf{0.0136}\\
     \hline
   \end{tabular}}
    \vspace{-0.15in}
   \caption{L2-CD ($\times100$) comparison under ShapeNet.}
   \label{table:t10}
    \vspace{-0.15in}
\end{table}

\begin{table}[tb]
\centering
\resizebox{\linewidth}{!}{
    \begin{tabular}{c|c|c|c|c|c|c|c|c|c|c|c}  
     \hline
        Class& PSR& DMC & BPA & ATLAS &DMC&DSDF&MeshP&LIG&IMNET&NP&Ours \\  
     \hline
        Display& 0.889& 0.842 & 0.952 & 0.828 &0.882&0.932& 0.974 & 0.926 &0.574&0.964&\textbf{0.9775}\\
        Lamp &0.876&0.872&0.951&0.593&0.725&0.864&\textbf{0.963}&0.882&0.592&0.930&0.9450\\
        Airplane&0.848&0.835&0.926&0.737&0.716&0.872&\textbf{0.955}&0.817&0.550&0.947&0.9490\\
        Cabinet&0.880&0.827&0.836&0.682&0.845&0.872&0.957&0.948&0.700&0.930&\textbf{0.9600}\\
        Vessel&0.861&0.831&0.917&0.671&0.706&0.841&0.953&0.847&0.574&0.941&\textbf{0.9546}\\
        Table&0.833&0.809&0.919&0.783&0.831&0.901&\textbf{0.962}&0.936&0.702&0.908&0.9595\\
        Chair&0.850&0.818&0.938&0.638&0.794&0.886&\textbf{0.962}&0.920&0.820&0.937&0.9580\\
        Sofa&0.892&0.851&0.940&0.633&0.850&0.906&\textbf{0.971}&0.944&0.818&0.951&0.9680\\
     \hline
     Mean&0.866&0.836&0.923&0.695&0.794&0.884&\textbf{0.962}&0.903&0.666&0.939&0.9590\\
     \hline
   \end{tabular}}
    \vspace{-0.15in}
   \caption{Normal consistency comparison under ShapeNet.}
   \label{table:t11}
    \vspace{-0.15in}
\end{table}

\begin{table}[tb]
\centering
\resizebox{\linewidth}{!}{
    \begin{tabular}{c|c|c|c|c|c|c|c|c|c|c|c|c|c}  
     \hline
        Class& PSR& DMC & BPA & ATLAS &DMC&DSDF& DGP &MeshP&NUD&LIG&IMNET&NP&Ours \\  
     \hline
        Display&0.468& 0.495& 0.834& 0.071& 0.108& 0.632& 0.417& 0.903& 0.903&0.551&0.601&0.989&\textbf{0.9939}\\
        Lamp& 0.455 &0.518& 0.826& 0.029 &0.047& 0.268& 0.405& 0.855&0.888&0.624&0.836&0.891&\textbf{0.9382}\\
        Airplane&0.415 &0.442& 0.788& 0.070& 0.050& 0.350& 0.249 &0.844&0.872&0.564&0.698&\textbf{0.996}&0.9942\\
        Cabinet&0.392 &0.392& 0.553& 0.077& 0.154 &0.573& 0.513& 0.860&0.950&0.733&0.343&0.980&\textbf{0.9888}\\
        Vessel&0.415& 0.466& 0.789& 0.058& 0.055& 0.323& 0.387& 0.862&0.883&0.467&0.147&0.985&\textbf{0.9935}\\
        Table&0.233& 0.287& 0.772& 0.080 &0.095& 0.577& 0.307& 0.880& 0.908&0.844&0.425&0.922&\textbf{0.9969}\\
        Chair&0.382 &0.433& 0.802& 0.050& 0.088& 0.447& 0.481& 0.875& 0.913&0.710&0.181&0.954&\textbf{0.9970}\\
        Sofa&0.499& 0.535& 0.786& 0.058& 0.129& 0.577& 0.638& 0.895&0.945&0.822&0.199&0.968&\textbf{0.9943}\\
     \hline
     Mean&0.407 &0.446 &0.769& 0.062 &0.091& 0.468& 0.425& 0.872&0.908&0.664&0.429&0.961&\textbf{0.9871}\\
     \hline
   \end{tabular}}
   \vspace{-0.15in}
   \caption{F-score($\mu$) comparison under ShapeNet.}
   \label{table:st12}
   \vspace{-0.15in}
\end{table}

\begin{table}[tb]
\centering
\resizebox{\linewidth}{!}{
    \begin{tabular}{c|c|c|c|c|c|c|c|c|c|c|c}  
     \hline
        Class& PSR& DMC & BPA & ATLAS &DMC&DSDF& DGP &MeshP&NUD&NP&Ours \\  
     \hline
        Display&0.666& 0.669& 0.929& 0.179& 0.246& 0.787& 0.607& 0.975&0.944&0.991&\textbf{0.9958}\\
        Lamp&0.648& 0.681& 0.934& 0.077& 0.113& 0.478& 0.662& \textbf{0.951}&0.945&0.924&0.9402\\
        Airplane&0.619& 0.639& 0.914& 0.179 &0.289& 0.566& 0.515& 0.946&0.944&0.997&\textbf{0.9972}\\
        Cabinet&0.598& 0.591& 0.706& 0.195& 0.128& 0.694& 0.738& 0.946&0.980&0.989&\textbf{0.9939}\\
        Vessel&0.633& 0.647& 0.906& 0.153& 0.120& 0.509& 0.648& 0.956&0.945&0.990&\textbf{0.9958}\\
        Table&0.442& 0.462& 0.886& 0.195& 0.221& 0.743& 0.494& 0.963&0.922&0.973&\textbf{0.9985}\\
        Chair&0.617& 0.615& 0.913& 0.134& 0.345& 0.665 & 0.693& 0.964&0.954&0.969&\textbf{0.9991}\\
        Sofa&0.725& 0.708& 0.895& 0.153& 0.208& 0.734& 0.834& 0.972&0.968&0.974&\textbf{0.9987}\\
     \hline
     Mean&0.618& 0.626& 0.885& 0.158& 0.209& 0.647& 0.649 &0.959&0.950&0.976&\textbf{0.9899}\\
     \hline
   \end{tabular}}
   \vspace{-0.15in}
   \caption{F-score($2\mu$) comparison under ShapeNet.}
   \label{table:t13}
   \vspace{-0.3in}
\end{table}

Moreover, we also compare with other methods which reported a reconstruction accuracy over the subset under ShapeNet in terms of L1-CD in Tab.~\ref{table:NOX2}. These methods include 3DR2~\cite{ChoyXGCS16}, PSGN~\cite{FanSG17}, DMC~\cite{Liao2018CVPR}, Occupancy Network (OccNet)~\cite{MeschederNetworks}, SSRNet~\cite{Mi_2020_CVPR}, DDT~\cite{luo2021deepdt}, and NP~\cite{Zhizhong2021icml}. We also report our accuracy under each involved class at the bottom of the Tab.~\ref{table:NOX2}. This comparison further demonstrates our ability to reconstruct surfaces at high accuracy.

\begin{figure}[tb]
  \centering
   \includegraphics[width=\linewidth]{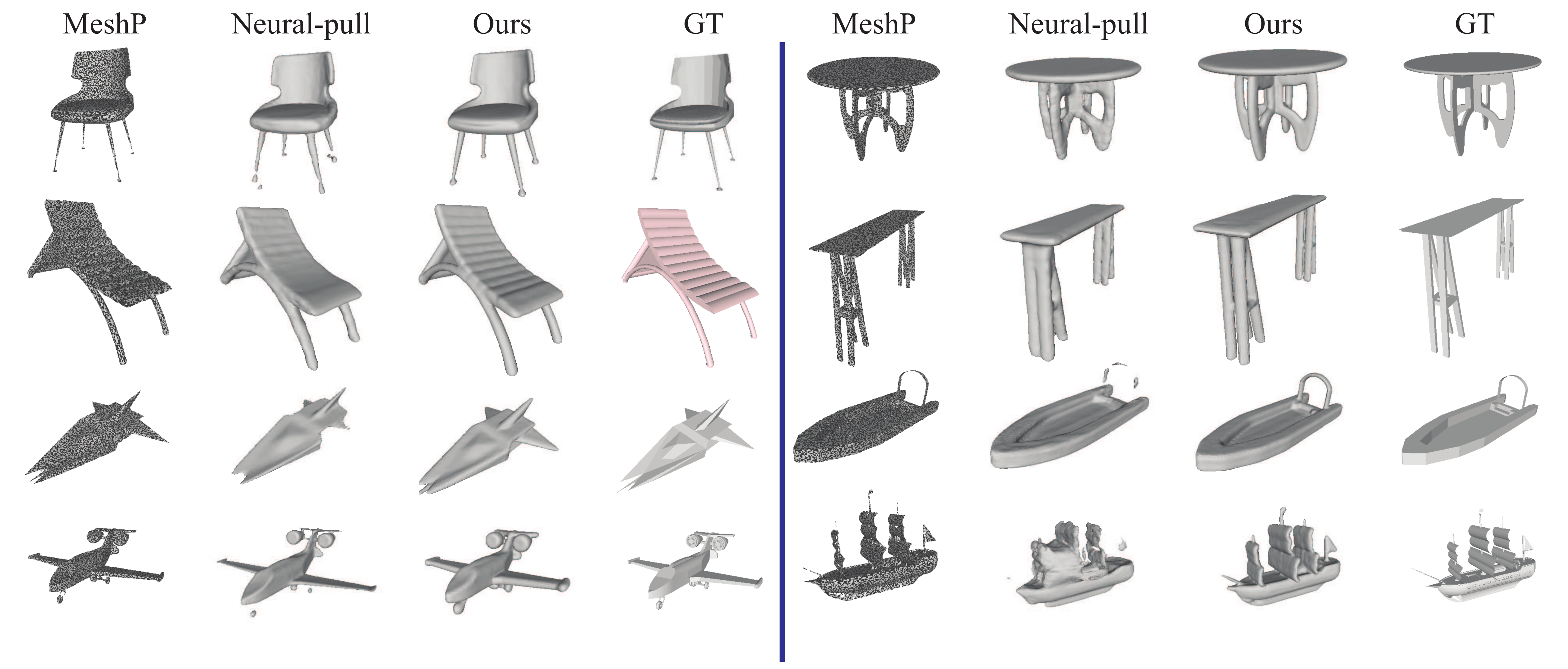}
  %
  %
  \vspace{-0.3in}
\caption{\label{fig:ComparShapeNet}Visual comparison under ShapeNet dataset.}
\vspace{-0.18in}
\end{figure}

We visually compare with P2S and NP in Fig.~\ref{fig:ComparShapeNet}. We present more accurate geometry on the complete surface while P2S fails to reconstruct the complete surface.

\begin{table}[h]
\centering
\resizebox{\linewidth}{!}{
    \begin{tabular}{c|c|c|c|c|c|c|c}  
     \hline
          3DR2 & PSGN & DMC& OccNet& SSRNet& DDT & NP& Ours\\   
     \hline
       0.169&0.202&0.117&0.079&0.024&0.020&0.011 & \textbf{0.0077}\\ 
     \hline
     \hline
       Display&Lamp&Airplane&Cabinet&Vessel&Table&Chair&Sofa\\
     \hline
       0.0073&0.0082&0.0057&0.0081&0.0071&0.0083&0.0080&0.0088\\
     \hline
   \end{tabular}}
   \vspace{-0.15in}
   \caption{Reconstruction accuracy in terms of L1-CD.}
   \label{table:NOX2}
   \vspace{-0.15in}
\end{table}

\begin{figure}[tb]
  \centering
   \includegraphics[width=\linewidth]{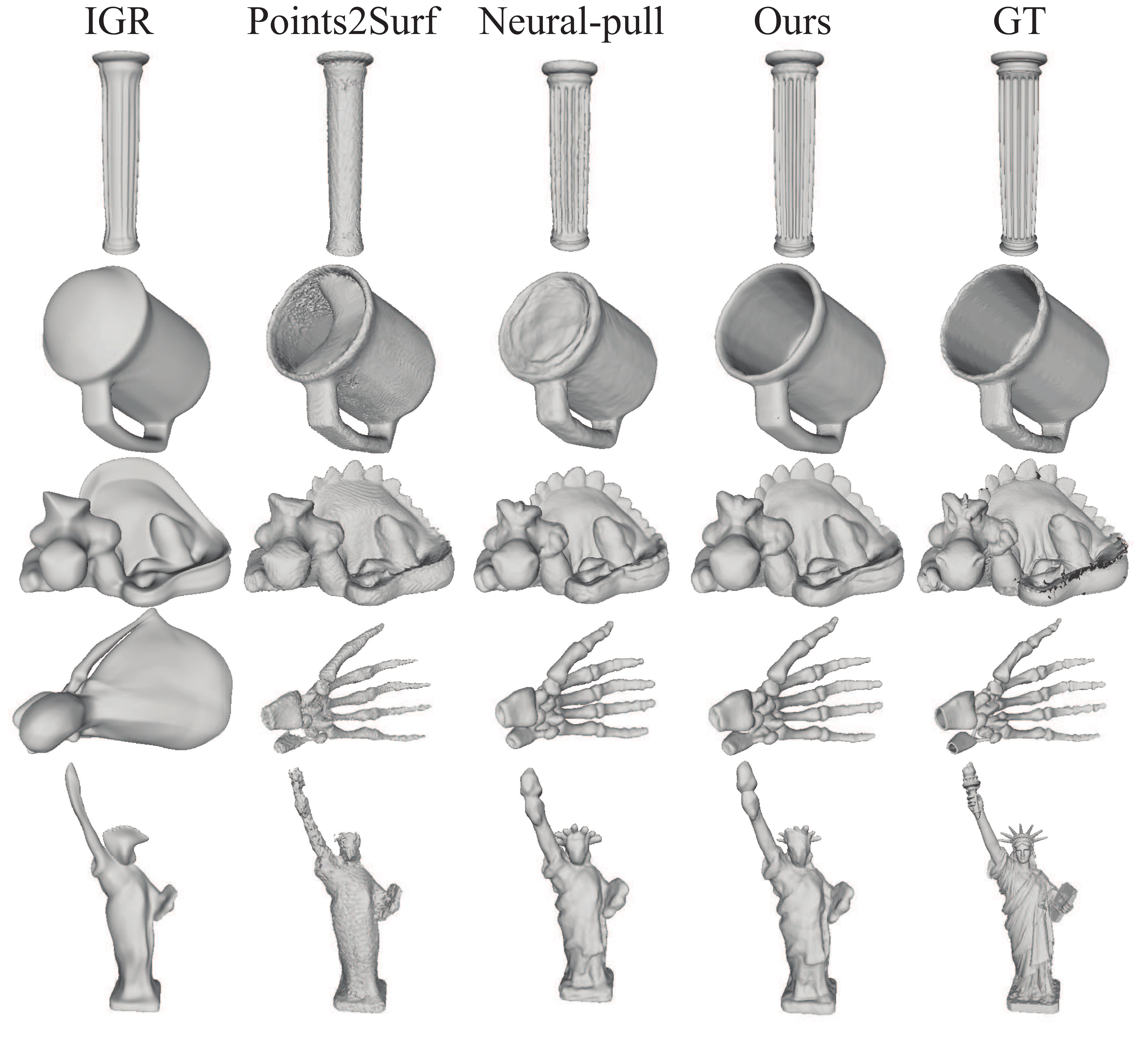}
  %
  %
  \vspace{-0.3in}
\caption{\label{fig:ComparFamous}Visual comparison under FAMOUS dataset.}
\vspace{-0.18in}
\end{figure}

\begin{figure}[tb]
  \centering
   \includegraphics[width=\linewidth]{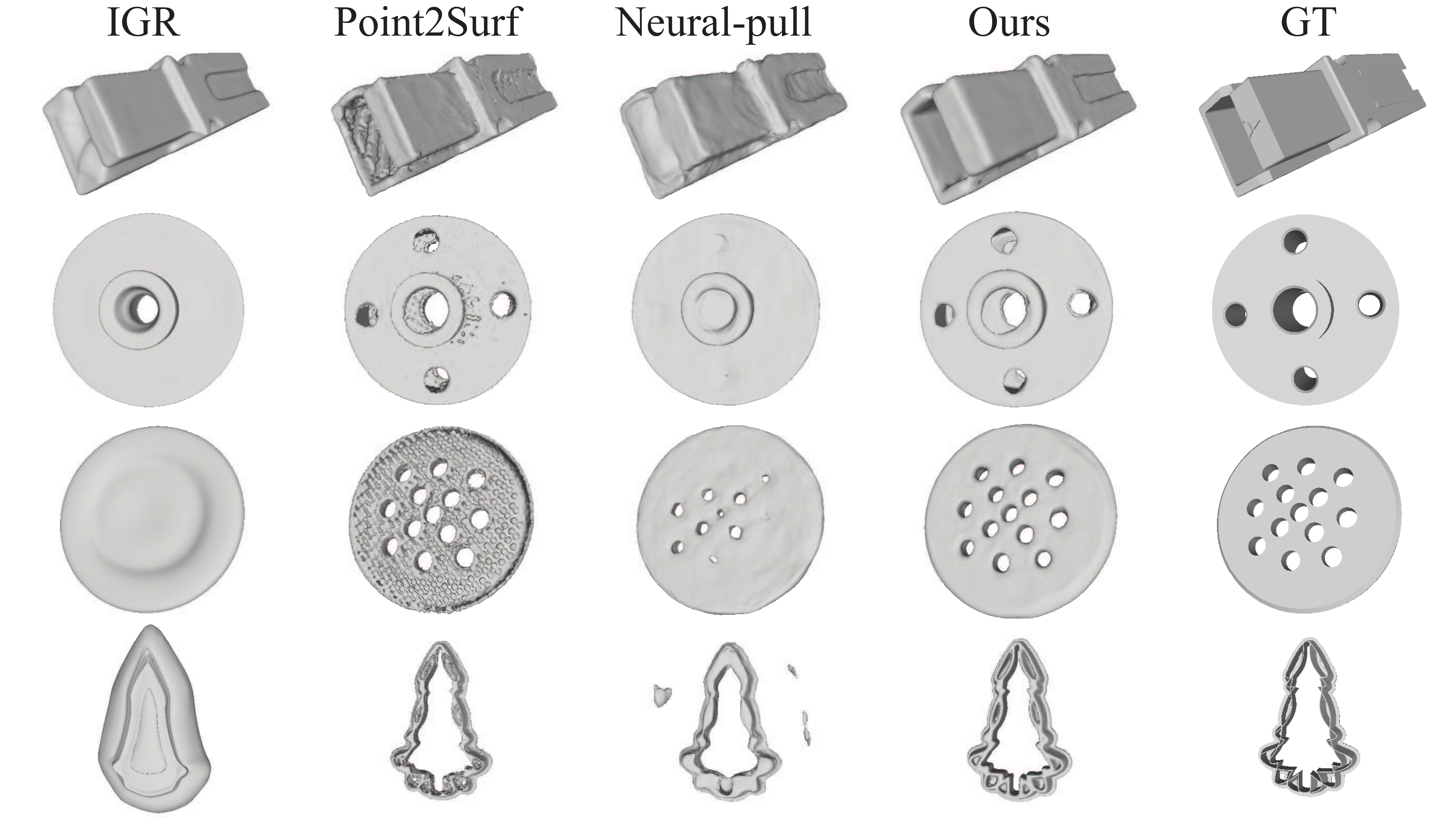}
  %
  %
  \vspace{-0.3in}
\caption{\label{fig:ABC}Visual comparison under ABC dataset.}
\vspace{-0.28in}
\end{figure}

\noindent\textbf{Evaluation under ABC and FAMOUS. }
Tab.~\ref{table:NOX1} reports the comparison under ABC and FAMOUS dataset with DeepSDF (DSDF)~\cite{Park_2019_CVPR}, AtlasNet (ATLAS)~\cite{Groueix_2018_CVPR}, PSR~\cite{journals/tog/KazhdanH13}, Points2Surf (P2S)~\cite{ErlerEtAl:Points2Surf:ECCV:2020}, IGR~\cite{gropp2020implicit}, Neural-Pull (NP)~\cite{Zhizhong2021icml} and IMLS~\cite{Liu2021MLS}. The numerical comparison shows that our method significantly outperforms the other methods. We also highlight our advantage by visually comparing with IGR, P2S, and NP under FAMOUS in Fig.~\ref{fig:ComparFamous} and under ABC in Fig.~\ref{fig:ABC}, where our reconstruction presents more geometry details with arbitrary topology.

We also evaluate our method under some variants of ABC and FAMOUS by adding different noise levels or changing the point density, which is released by P2S~\cite{ErlerEtAl:Points2Surf:ECCV:2020}. The L2-CD comparison in Tab.~\ref{table:NOX3noise} demonstrates that our method is also good at resisting dramatic noise and density changes, and still achieves the best performance compared to the others.\vspace{-0.13in}

\begin{table}[h]
\centering
\resizebox{\linewidth}{!}{
    \begin{tabular}{c|c|c|c|c|c|c|c}  
     \hline
          Dataset& DSDF & ATLAS & PSR& P2S& IGR & NP& Ours\\   
     \hline
       ABC& 8.41 & 4.69 & 2.49& 1.80& 0.51 & 0.48 &\textbf{0.200}\\ 
       FAMOUS &10.08&4.69&1.67&1.41&1.65&0.22 &\textbf{0.044}\\
     \hline
     Mean&9.25&4.69&2.08&1.61&1.08&0.35&\textbf{0.122}\\
     \hline
   \end{tabular}}
   \vspace{-0.15in}
   \caption{Reconstruction accuracy in terms of L2-CD ($\times100$).}
   \vspace{-0.28in}
   \label{table:NOX1}
\end{table}

\begin{table}[h]
\centering
\resizebox{\linewidth}{!}{
    \begin{tabular}{c|c|c|c|c|c|c|c}  
     \hline
          Dataset& DSDF & ATLAS & PSR & P2S& NP&IMLS&Ours\\   
     \hline
       ABC var-noise& 12.51 & 4.04 & 3.29& 2.14&0.72 & 0.566&\textbf{0.488} \\ 
       ABC max-noise& 11.34 & 4.47 & 3.89& 2.76&1.24& 0.675&\textbf{0.571} \\ 
       F-med-noise &9.89&4.54&1.80&1.51&0.28&0.798&\textbf{0.071}\\
       F-max-noise &13.17&4.14&3.41&2.52&0.31&0.387&\textbf{0.298}\\
       \hline
       F-Sparse&10.41&4.91&2.17&1.93&0.84&-&\textbf{0.083}\\
       F-Dense&9.49&4.35&1.60&1.33&0.22&-&\textbf{0.087}\\
     \hline
     Mean &11.73&4.30&3.10&2.23&0.60&-&\textbf{0.266}\\
     \hline
   \end{tabular}}
   \vspace{-0.15in}
   \caption{Noise and density in terms of L2-CD ($\times100$).}
   \vspace{-0.25in}
   \label{table:NOX3noise}
\end{table}

\subsection{Surface Reconstruction for Scenes}
\begin{figure*}[tb]
  \centering
   \includegraphics[width=\linewidth]{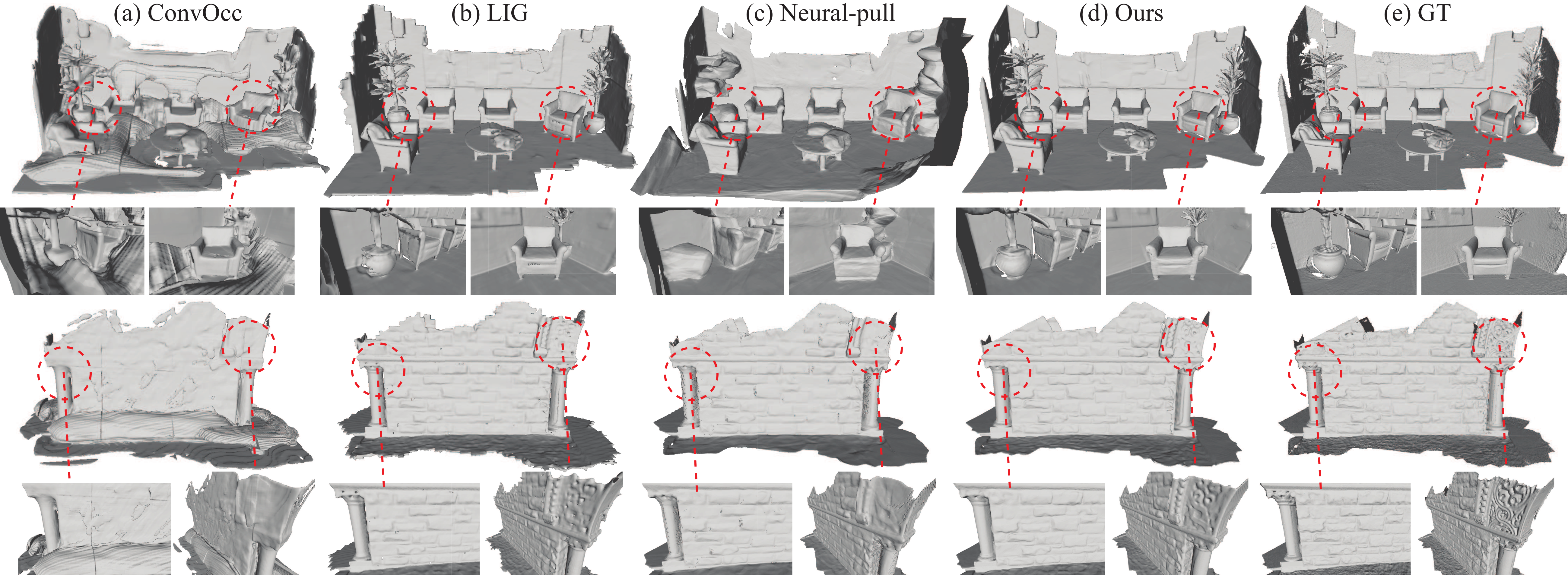}
  %
  %
  \vspace{-0.3in}
\caption{\label{fig:eccv}Visual comparison with the state-of-the-art under 3D Scene dataset.}
\vspace{-0.15in}
\end{figure*}

\noindent\textbf{Evaluation under 3D Scene. }We first evaluate our method by comparing with MPU~\cite{OhtakeBATS03}, Convolutional OccNet (ConvOcc)~\cite{Peng2020ECCV}, Local Implicit Grid (LIG)~\cite{jiang2020lig}, Deep Local Shape (DeepLS)~\cite{DBLP:conf/eccv/ChabraLISSLN20}, and Neural-Pull (NP)~\cite{Zhizhong2021icml} under five scenes in the 3D Scene dataset in Tab.~\ref{table:t12}. We use the official code of MPU and NP to produce their results, while using the trained ConvOcc and LIG from the author and normals of point clouds to report their results, where we do not use the post processing in LIG for fair comparison. Tab.~\ref{table:t12} shows that our method can achieve much higher accuracy than these state-of-the-art methods in terms of different metrics, where we also do not require normals as LIG and DeepLS. The improvement over the-state-of-the-art is further demonstrated by the visual comparison in Fig.~\ref{fig:eccv}.

\begin{table*}[tb]
\centering
\resizebox{\linewidth}{!}{
    \begin{tabular}{c|c|c|c|c||c|c|c|c||c|c|c|c||c|c|c|c||c|c|c|c}
     \hline

        &\multicolumn{4}{c||}{Burghers}&\multicolumn{4}{c||}{Lounge}&\multicolumn{4}{c||}{Copyroom}&\multicolumn{4}{c||}{Stonewall}&\multicolumn{4}{c}{Totempole}\\
        \hline
        &L2CD&L1CD&Norm&Error&L2CD&L1CD&Norm&Error&L2CD&L1CD&Norm&Error&L2CD&L1CD&Norm&Error&L2CD&L1CD&Norm&Error\\
     \hline
     MPU&775.04&0.456&0.720&894.95&203.87&0.206&0.817&203.87&29.18&0.062&0.832&110.36&680.86&0.428&0.800&486.70&1652.25&0.671&0.763&1328.80\\
     ConvOcc&26.69&0.077&0.865&354.60&8.68&0.042&0.857&73.20&10.99&0.045&0.848&79.68&19.12&0.066&0.866&120.61&1.16&0.016&\textbf{0.925}&21.56\\
     LIG&0.839&0.018&0.904&28.70&0.789&0.017&0.910&28.20&0.906&0.018&0.910&30.50&1.08&0.020&\textbf{0.928}&33.65&1.37&0.023&0.917&38.90\\
     DeepLS&-&-&-&\textbf{5.74}&-&-&-&7.38&-&-&-&10.09&-&-&-&6.45&-&-&-&8.97\\
     NP&1.76&0.010&0.883&11.23&39.71&0.059&0.857&98.03&0.51&0.011&0.884&8.76&0.063&0.007&0.868&6.84&0.19&0.010&0.765&10.21\\
     \hline
     Ours&\textbf{0.267}&\textbf{0.008}&\textbf{0.914}&9.28&\textbf{0.061}&\textbf{0.006}&\textbf{0.928}&\textbf{6.76}&\textbf{0.076}&\textbf{0.007}&\textbf{0.918}&\textbf{7.76}&\textbf{0.061}&\textbf{0.0065}&0.888&\textbf{6.33}&\textbf{0.10}&\textbf{0.008}&0.784&\textbf{8.36}\\
     \hline
   \end{tabular}}
   \vspace{-0.15in}
   \caption{Surface reconstruction under 3D Scene.L2-CD$\times 1000$. Norm is short for normal consistency. The unit of error is mm.}
   \label{table:t12}
   \vspace{-0.25in}
\end{table*}

\begin{figure*}[tb]
  \centering
   \includegraphics[width=\linewidth]{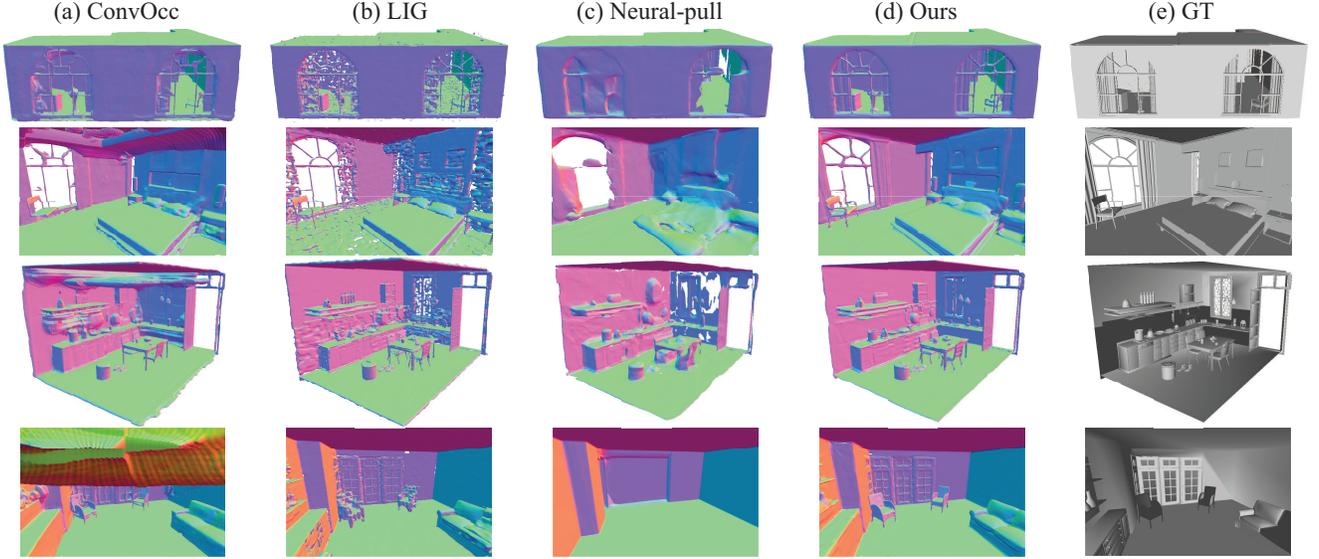}
  %
  %
  \vspace{-0.3in}
\caption{\label{fig:cvpr}Visual comparison with the state-of-the-art under SceneNet dataset, where normal maps are also shown.}
\vspace{-0.25in}
\end{figure*}

\noindent\textbf{Evaluation under SceneNet. }We compare our method with ConvOcc~\cite{Peng2020ECCV}, LIG~\cite{jiang2020lig}, and NP~\cite{Zhizhong2021icml} under 5 classes in SceneNet. We produce the results of NP by training it using its code, while using the trained model of LIG to produce their results, where we do not leverage the post processing in LIG for fair comparison. The results in each of five classes in Tab.~\ref{table:t12lig} show that our method achieves the best performances under different input point densities. Our visual comparison in Fig.~\ref{fig:cvpr} further shows that our method can reconstruct more detailed surfaces in complex scenes.

\noindent\textbf{Reconstructions for Real Scan. }We also show surface reconstruction comparison for a real scanned scene in our video and text supplementary.

\begin{table*}[tb]
\centering
\resizebox{\linewidth}{!}{
    \begin{tabular}{c|c|c|c|c||c|c|c||c|c|c||c|c|c||c|c|c||c|c|c}
     \hline

        &&\multicolumn{3}{c||}{Livingroom}&\multicolumn{3}{c||}{Bathroom}&\multicolumn{3}{c||}{Bedroom}&\multicolumn{3}{c||}{Kitchen}&\multicolumn{3}{c||}{Office}&\multicolumn{3}{c}{Mean}\\
        \hline
        &&L1CD&Norm&FScore&L1CD&Norm&FScore&L1CD&Norm&FScore&L1CD&Norm&FScore&L1CD&Norm&FScore&L1CD&Norm&FScore\\
        \hline
        \multirow{3}{*}{\rotatebox{90}{20/$m^2$}}&LIG&0.032&0.719&0.790&\textbf{0.030}&0.737&\textbf{0.807}&0.029&0.735&0.818&\textbf{0.029}&0.727&0.817&0.033&0.737&0.805&0.030&0.730&0.808\\
        &NP&0.068&0.827&0.718&0.072&0.716&0.658&0.044&0.782&0.740&0.069&0.720&0.689&0.066&0.834&0.663&0.037&0.776&0.693\\
        \cline{2-20}
        &Ours&\textbf{0.027}&\textbf{0.835}&\textbf{0.856}&0.032&\textbf{0.749}&0.801&\textbf{0.028}&\textbf{0.800}&\textbf{0.842}&0.033&\textbf{0.737}&\textbf{0.826}&\textbf{0.029}&\textbf{0.861}&\textbf{0.829}&\textbf{0.029}&\textbf{0.796}&\textbf{0.831}\\
     \hline
        \multirow{3}{*}{\rotatebox{90}{100/$m^2$}}&LIG&0.019&\textbf{0.922}&0.919&0.018&\textbf{0.930}&0.915&0.017&\textbf{0.918}&0.920&0.016&\textbf{0.920}&0.936&\textbf{0.020}&0.910&\textbf{0.936}&0.018&\textbf{0.920}&0.925\\
        &NP&0.069&0.883&0.799&0.028&0.907&0.893&0.032&0.890&0.878&0.042&0.896&0.838&0.066&0.866&0.733&0.047&0.888&0.828\\
        \cline{2-20}
        &Ours&\textbf{0.018}&0.902&\textbf{0.953}&\textbf{0.016}&0.872&\textbf{0.950}&\textbf{0.014}&0.893&\textbf{0.957}&\textbf{0.015}&0.884&\textbf{0.945}&0.024&\textbf{0.916}&0.907&\textbf{0.017}&0.893&\textbf{0.943}\\
     \hline
        \multirow{3}{*}{\rotatebox{90}{500/$m^2$}}&LIG&0.019&0.910&0.919&0.017&0.924&0.914&0.017&0.915&0.926&0.017&0.916&0.937&\textbf{0.020}&0.907&\textbf{0.937}&0.018&0.914&0.925\\
        &NP&0.050&0.905&0.838&0.041&0.916&0.856&0.033&0.915&0.877&0.047&0.893&0.844&0.064&0.879&0.750&0.047&0.902&0.833\\
        \cline{2-20}
        &Ours&\textbf{0.017}&\textbf{0.938}&\textbf{0.969}&\textbf{0.016}&\textbf{0.943}&\textbf{0.979}&\textbf{0.016}&\textbf{0.946}&\textbf{0.976}&\textbf{0.016}&\textbf{0.943}&\textbf{0.968}&0.024&\textbf{0.927}&0.918&\textbf{0.017}&\textbf{0.939}&\textbf{0.962}\\
     \hline
        \multirow{3}{*}{\rotatebox{90}{1000/$m^2$}}&LIG&0.020&0.910&0.920&0.017&0.927&0.910&0.017&0.919&0.924&0.017&0.920&0.936&\textbf{0.020}&0.910&\textbf{0.936}&0.018&0.917&0.925\\
        &NP&0.088&0.881&0.801&0.036&0.912&0.860&0.034&0.905&0.876&0.049&0.900&0.825&0.062&0.879&0.729&0.054&0.895&0.818\\
        \cline{2-20}
        &Ours&\textbf{0.017}&\textbf{0.933}&\textbf{0.966}&\textbf{0.016}&\textbf{0.945}&\textbf{0.977}&\textbf{0.014}&\textbf{0.948}&\textbf{0.980}&\textbf{0.015}&\textbf{0.945}&\textbf{0.976}&0.024&\textbf{0.919}&0.925&\textbf{0.017}&\textbf{0.938}&\textbf{0.965}\\
     \hline
   \end{tabular}}
   \vspace{-0.15in}
   \caption{Surface reconstruction under SceneNet.}
   \label{table:t12lig}
   \vspace{-0.20in}
\end{table*}

\subsection{Analysis and Discussion}
We justify the effectiveness of each element in our network and explore the effect of some important parameters on the performance under the ABC dataset in terms of L2-CD and normal consistency.

\noindent\textbf{Ablation Studies. }We report ablation studies in Tab.~\ref{table:NOX11}. We first highlight the effectiveness of the predicted shift $\Delta \bm{q}$ by removing $\Delta \bm{q}$ from the network. We first try to directly use the query $\bm{q}_g$ from the global coordinate system as $\bm{q}_l'$. We found that the performance degenerates dramatically, as shown by ``No $\Delta \bm{q}$''. Then, we push the neural query network to predict $\bm{q}_l'$ directly. But the performance still goes down, as shown by ``Direct $\bm{q}_l'$''. These two results demonstrate the importance of $\Delta \bm{q}$ in the learning.

\begin{wrapfigure}{r}{0.5\linewidth}
\vspace{-0.15in}
\includegraphics[width=\linewidth]{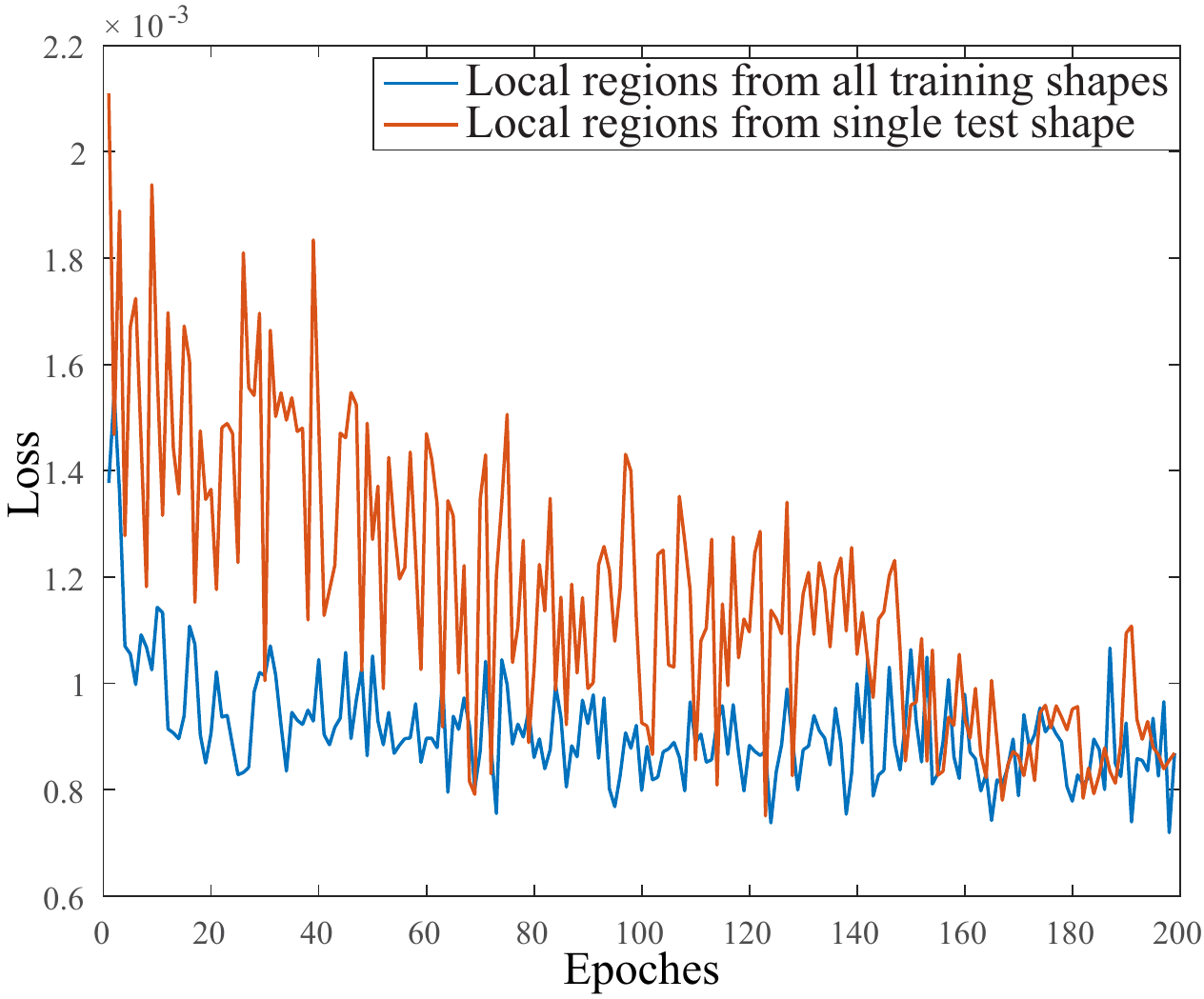}
\vspace{-0.25in}
\caption{\label{fig:1} Loss with the prior learned from different regions.}
\vspace{-0.15in}
\end{wrapfigure}

Moreover, we highlight the effectiveness of the predicted condition $\bm{f}_l'$ by removing it from the output of the neural query network. We first leverage autodecoding similar as DeepSDF~\cite{Park_2019_CVPR} to learn $\bm{f}_l'$. The result of ``No $\bm{f}_l'$'' shows that this does not work well with the learnable $\Delta \bm{q}$. Then, we try to use $\bm{f}_l$ from the trained PointNet to replace $\bm{f}_l'$, but the result of ``No $\bm{f}_l'$+$\bm{f}_l$'' gets worse neither. These experiments show that the learnable condition $\bm{f}_l'$ is only effective when it is optimized together with its corresponding query $\bm{q}_l'$.

\begin{table}[h]
\centering
  \vspace{-0.15in}
\resizebox{\linewidth}{!}{
    \begin{tabular}{c|c|c|c|c|c}  
     \hline
          & No $\Delta \bm{q}$ & Direct $\bm{q}_l'$ & No $\bm{f}_l'$& No $\bm{f}_l'$+$\bm{f}_l$& Ours\\   
     \hline
       L2-CD& 3.13&2.78&4.03&4.21&\textbf{2.090}\\ 
       Normal& 0.924&0.935&0.908&0.901&\textbf{0.945}\\
       \hline
   \end{tabular}}
   \vspace{-0.15in}
   \caption{Ablation studies under ABC. L2-CD$\times 1000$.}
   \vspace{-0.2in}
   \label{table:NOX11}
\end{table}

\noindent\textbf{Specializing Context Prior. }We found that specializing the local context prior into the predictive context prior also plays an important role to reconstruct surfaces in high accuracy. We first highlight the importance of the local context prior by removing the training procedure shown in Fig.~\ref{fig:Overview} (a). We overfit the global point cloud $\bm{G}$ by simultaneously optimizing the parameters $\bm{\theta}_3$ in the neural query network and $\bm{\theta}_2$ in the neural implicit network shown in Fig.~\ref{fig:Overview} (b). The result of ``No prior'' in Tab.~\ref{table:NOX12} shows that the performance significantly drops compared to ``Our specializing''. Moreover, even with the learned local context prior as an initialization, if we tune $\bm{\theta}_2$ and $\bm{\theta}_3$ simultaneously in Fig.~\ref{fig:Overview} (b), the result of ``Tune $\bm{\theta}_2$+$\bm{\theta}_3$'' is still not satisfactory. These experiments demonstrate the importance of the specializing in leveraging the learned prior.

\begin{table}[h]
\centering
  \vspace{-0.15in}
    \begin{tabular}{c|c|c|c}  
     \hline
          & No Prior& Tune $\bm{\theta}_2$+$\bm{\theta}_3$&Our specializing\\   
     \hline
       L2-CD&4.04&3.60&\textbf{2.09}\\
       Normal&0.9200&0.9250&\textbf{0.9446}\\
       \hline
   \end{tabular}
   \vspace{-0.15in}
   \caption{Effect of specializing under ABC. L2-CD$\times 1000$.}
   \vspace{-0.15in}
   \label{table:NOX12}
\end{table}

\noindent\textbf{Normalizing Local Regions. }We found the normalization of local regions $\bm{t}_i$ in $\mathbf{T}$ slightly affects the learning of the local context prior. As we mentioned before, we normalize $\bm{t}_i$ by centering and scaling it in the local coordinate system.
We report the effect of centering and scaling on the performance in Tab.~\ref{table:NOX131}, which shows that both centering and scaling contribute to the increase of performance.

\begin{table}[h]
\centering
  \vspace{-0.15in}
\resizebox{\linewidth}{!}{
    \begin{tabular}{c|c|c|c|c}  
     \hline
          & No normalization&Only centering&Only scale& Ours \\   
     \hline
       L2-CD&2.83&2.13&2.67&\textbf{2.09}\\
       Normal&0.9360&0.9410&0.9380&\textbf{0.9446}\\
       \hline
   \end{tabular}}
   \vspace{-0.15in}
   \caption{Effect of normalization under ABC. L2-CD$\times 1000$.}
   \vspace{-0.15in}
   \label{table:NOX131}
\end{table}

\noindent\textbf{Obtaining Local Regions $\bm{t}_i$. }We also explore how the size of local regions $\bm{t}_i$ affects the learning of the local context prior. We try to split each point cloud in the training set into different numbers of parts, such as $\{0^3,4^3,6^3,8^3\}$, and then, use each set of parts to learn the local context prior in Fig.~\ref{fig:Overview} (a). The comparison shown in Tab.~\ref{table:NOX13} demonstrates that it is hard to learn the prior well if the size of $\bm{t}_i$ is too large, such as ``$0^3$'' and ``$4^3$'', since $\bm{t}_i$ is too complex to learn. While it is also hard to learn some meaningful prior if the size of $\bm{t}_i$ is too small, such as ``$8^3$''. In addition, we found that the overlap between neighboring local regions does not contribute to the performance increasing under $6^3$, such as ``Lap'' and ``Self''. We also explore whether we can learn a more meaningful prior by using patch-wise $\bm{t}_i$ in training. We form each $\bm{t}_i$ using 1000, 2000, or 4000 neighbors in terms of geodesic distance. The results of ``G1'', ``G2'' and ``G4'' show that patch-wise regions $\bm{t}_i$ do not work better than the part-wise $\bm{t}_i$ that we are using.

We highlight our advantage by learning the local context prior using local regions $\bm{t}_i$ merely from the reconstruction target $\bm{G}$. Although ``Self'' is obtained with much fewer training regions, it achieves almost the same result as ``$6^3$'' which is obtained by learning the local context prior from all local regions across different training shapes. This advantage comes from our ability of flexibly searching over the whole prior space, which alleviates the necessity of learning a high quality local context prior. However, the optimization
can converge faster if more local regions $\bm{t}_i$ are used in learning
as shown by the loss curve comparison in Fig.~\ref{fig:1}.\vspace{-0.13in}

 \begin{table}[h]
\centering
\resizebox{\linewidth}{!}{
    \begin{tabular}{c|c|c|c|c||c|c||c|c|c}  
     \hline
          & $0^3$&$4^3$&$6^3$&$8^3$& Lap & Self &G1& G2& G4\\   
     \hline
       L2-CD&12.06&4.43&\textbf{2.09}&2.58&2.09&2.09&2.9&2.3&2.5\\
       Normal&0.904&0.922&\textbf{0.945}&0.940&0.944&0.942&0.939&0.941&0.940\\
       \hline
   \end{tabular}}
   \vspace{-0.15in}
   \caption{Effect of region size under ABC. L2-CD$\times 1000$.}
   \vspace{-0.2in}
   \label{table:NOX13}
\end{table}

\noindent\textbf{Limitation. }Although we achieve high reconstruction accuracy, we require further optimization during testing. This takes more time than methods~\cite{ErlerEtAl:Points2Surf:ECCV:2020} leveraging pretrained models for inference.

\section{Conclusion}
We propose to reconstruct surfaces from point clouds by learning implicit functions as a predictive context prior. Our method successfully specializes the learned local context prior into predictive context prior for a specific point cloud, which effectively searches the reconstruction prior across the whole prior space without focusing on some specific locations. This advantage significantly increases our ability of leveraging the learned prior, which makes the learned prior generalize to as many unseen target regions as possible. Our idea is justified by our experimental results which outperform the state-of-the-art in terms of various metrics under widely used benchmarks.

{\small
\bibliographystyle{ieee_fullname}
\bibliography{PaperForReview}
}

\end{document}